\relax
\documentclass[letterpaper]{article} 
\usepackage{arxiv}  
\usepackage{times}  
\usepackage{helvet}  
\usepackage{courier}  
\usepackage[hyphens]{url}  
\usepackage{graphicx} 
\usepackage{natbib}  
\usepackage{caption} 
\DeclareCaptionStyle{ruled}{labelfont=normalfont,labelsep=colon,strut=off} 
\usepackage{algorithm}
\usepackage{algorithmic}

\usepackage{newfloat}
\usepackage{listings}
\floatstyle{ruled}
\newfloat{listing}{tb}{lst}{}
\floatname{listing}{Listing}
\title{Will My Robot Achieve My Goals? Predicting the Probability that an MDP Policy Reaches a User-Specified Behavior Target}
\author {
    Alexander Guyer\textsuperscript{1} \textmd{and} Thomas G. Dietterich\textsuperscript{2}\\
    Collaborative Robotics and Intelligent Systems (CoRIS) Institute\\
    Oregon State University\\
    Corvallis, OR 97331 USA\\
    \texttt{guyera@oregonstate.edu}\textsuperscript{1}, \texttt{tgd@cs.orst.edu}\textsuperscript{2}\\
}

\renewcommand{\shorttitle}{Will My Robot Achieve My Goals?}

\usepackage{amsmath}
\usepackage{amsthm}
\usepackage{bbm}

\newtheorem{theorem}{Theorem}
\newtheorem{lemma}{Lemma}
\newtheorem{definition}{Definition}
\let\vec\mathbf
\newcommand{\refeq}[1]{(\ref{eq:#1})}
\newcommand{\refthm}[1]{Theorem \ref{thm:#1}}
\newcommand{\reflemma}[1]{Lemma \ref{lemma:#1}}
\newcommand{\reffig}[1]{Figure \ref{fig:#1}}

\begin{document}

\maketitle

\begin{abstract}
As an autonomous system performs a task, it should maintain a calibrated estimate of the probability that it will achieve the user's goal. If that probability falls below some desired level, it should alert the user so that appropriate interventions can be made. This paper considers settings where the user's goal is specified as a target interval for a real-valued performance summary, such as the cumulative reward, measured at a fixed horizon $H$. At each time $t \in \{0, \ldots, H-1\}$, our method produces a calibrated estimate of the probability that the final cumulative reward will fall within a user-specified target interval $[y^-,y^+].$ Using this estimate, the autonomous system can raise an alarm if the probability drops below a specified threshold. We compute the probability estimates by inverting conformal prediction. Our starting point is the Conformalized Quantile Regression (CQR) method of Romano et al., which applies split-conformal prediction to the results of quantile regression. CQR is not invertible, but by using the conditional cumulative distribution function (CDF) as the non-conformity measure, we show how to obtain an invertible modification that we call \textbf{P}robability-space \textbf{C}onformalized \textbf{Q}uantile \textbf{R}egression (PCQR). Like CQR, PCQR produces well-calibrated conditional prediction intervals with finite-sample marginal guarantees. By inverting PCQR, we obtain guarantees for the probability that the cumulative reward of an autonomous system will fall below a threshold sampled from the marginal distribution of the response variable (i.e., a calibrated CDF estimate)that we employ to predict coverage probabilities for user-specified target intervals. Experiments on two domains confirm that these probabilities are well-calibrated.
\end{abstract}

\section{Introduction}
\label{sec:introduction}

\noindent Consider an agent executing a fixed policy $\pi$ in a Markov Decision Process. The user specifies some real-valued behavior quantity $b_t$ that summarizes the behavior of the agent from time $0$ through time $t$. For example, in this paper, $b_t = r_0 + \cdots + r_{t-1}$ is the sum of the rewards received over the first $t$ steps of the trajectory. The user also specifies a target interval $[y^-,y^+]$. At each time step $t$, the agent should produce a well-calibrated estimate of the probability, $P[y^- \leq b_H \leq y^+ | s_t]$, that the behavior summary at time $H$ falls within the target interval given the current MDP state $s_t$. If this probability falls below a specified threshold (e.g., $0.99$), the agent should alert the user. 

This implements an ``autonomy contract'' in which the system acts autonomously but is obligated to report back to the user if the probability of success falls below a specified level.  We will consider two application domains in this paper: Starcraft 2 and invasive species management. In Starcraft 2, if the probability of hitting the target interval drops too low, the user might be able to provide additional resources or reinforcements. In the invasive species management problem, if the probability of achieving the target total reward is too small, the conservation manager could request an increase in the management budget. Importantly, these actions need to be taken as soon as the \textit{predicted probability} is too low rather than waiting until $t=H$, by which time it is too late to take corrective action. This autonomy contract can also be viewed as a module within a hierarchical MDP policy.

The problem of estimating $P[y^- \leq b_H \leq y^+ | s_t]$ could be easily solved if we could compute the \textit{conditional cumulative distribution function}, $F(b_H | s_t)$. This predicts the entire CDF of $b_H$ conditioned on the current MDP state $s_t$.  Given exact estimates, $P[y^- \leq b_H \leq y^+]=F(y^+|s_t) - F(y^-|s_t)$. Methods for estimating the CDF have been developed recently \citep{implicit_quantile_networks,regression_quantiles}, but they do not provide finite-sample accuracy guarantees. To obtain those, we extend the methodology of conformal prediction to invert Conformalized Quantile Regression \cite[CQR;][]{cqr}.

\section{Related work}

We begin with previous work on the ``forward problem'' of constructing calibrated prediction intervals for regression. Given a conditioning input $\vec{x}$ and some significance level $\delta$, predict an interval about the response variable $Y$ with a coverage probability of $1 - \delta$. This problem is similar to standard regression, but the target output is a $1 - \delta$ prediction interval rather than a point estimate of the conditional mean.  Given its similarity to standard regression, a natural solution is to modify standard regression methods to instead predict conditional quantiles. A prediction can then be computed as the interval,
\begin{equation}
    \label{eq:quantile_regression}
    QR(\vec{x}, \delta) = \left[\hat{Q}_{Y|\vec{x}}\left(\frac{\delta}{2}\right), \hat{Q}_{Y|\vec{x}}\left(1 - \frac{\delta}{2}\right)\right],
\end{equation}
\noindent where $\hat{Q}_{Y|\vec{x}}(\alpha)$ is the estimated $\alpha$-quantile of the response variable $Y$ conditioned on the feature vector $\vec{x}$. This strategy is referred to as quantile regression, and it first appeared in the form of a modification to linear regression \cite{regression_quantiles}. Newer quantile regression models exist today, such as quantile regression forests \cite{qrf} and quantile neural networks \cite{qnn}.

Quantile regression minimizes an appropriate loss function, such as the pinball loss \citep{regression_quantiles}, averaged over the training data set. This leads to compromises such that decreasing the loss for some $\vec{x}$ may lead to increased losses for other $\vec{x}$ values. Consequently, quantile regression intervals fail to achieve \textit{validity}---they are not guaranteed to cover the test data with at least probability $1 - \delta$. In practice, quantile regression intervals can suffer from significant undercoverage or overcoverage at test time \citep{cqr}.

To overcome the invalidity of quantile regression estimates, many modern solutions are derived instead from the technique of conformal prediction \citep{alg_learn_in_a_rand_world}. Conformal prediction is a distribution-free frequentist approach that provides prediction intervals with finite-sample guarantees for marginal coverage probability. The more computationally-attractive versions hold out a ``calibration set'' in a style referred to as \textit{split conformal prediction}. Conformal prediction strategies exist for classification and regression tasks. In the case of regression, conformal prediction typically involves regressing the conditional expectation, as in standard regression, and then forming a symmetric, constant-width prediction band around the point estimate,
\begin{equation}
    \label{eq:conformal_prediction}
    CP(\vec{x}, \delta) = [\hat{y} - c_{\delta}, \hat{y} + c_{\delta}],
\end{equation}
\noindent where $\hat{y}$ is the estimated expectation of $Y$ conditioned on $\vec{x}$, and $c_{\delta}$ is a \textit{conformity score} selected to achieve at least $1 - \delta$ coverage over the calibration set. Provided that the data points in the calibration set are exchangeable draws from a stationary distribution, the predicted intervals are well-calibrated and marginally valid even with a finite sample, meaning that they are guaranteed to cover the test data with at least probability $1 - \delta$ in expectation over future exchangeable draws from the same distribution.

Conformal prediction has seen many uses and modifications in the regression setting. For instance, Vovk et al.~\citeyearpar{cp_online} further applied conformal prediction to online linear regression. Lei et al.~\citeyearpar{distribution_free_prediction_sets, distribution_free_prediction_bands} combined conformal prediction with nonparametric smoothing to improve the asymptotic efficiency of the prediction intervals under many function classes. Other examples abound.

However, the constant-width property of standard conformal prediction intervals can be overly restrictive. The variance of a conditional random variable $Y$ is often heterogeneous (i.e., dependent on the value of the conditioning variable $\vec{x}$). For instance, time series data is often heteroskedastic with the variance increasing along with the horizon due to the accumulation of uncertainty over time. Constant-width conditional prediction intervals computed for data with heterogeneous variance tend to be \textit{inefficient}, meaning that they are wider than necessary.

In an effort to achieve more efficient intervals while retaining marginal validity, Romano et al.~introduced an elegant method known as Conformalized Quantile Regression \cite[CQR;][]{cqr}. CQR borrows techniques from both quantile regression and conformal prediction by applying a conformalized ``correction'' to the standard quantile regression interval. The resulting prediction is the corrected interval,
\begin{equation}
    \label{eq:conformalized_quantile_regression}
    CQR(\vec{x}, \delta) = \left[\hat{Q}_{Y|\vec{x}}\left(\frac{\delta}{2}\right) - c_{\delta}, \hat{Q}_{Y|\vec{x}}\left(1 - \frac{\delta}{2}\right) + c_{\delta}\right],
\end{equation}
\noindent where $\hat{Q}_{Y|\vec{x}}(\alpha)$ is the conditional $\alpha$-quantile estimate (the quantile regression prediction), and $c_{\delta}$ is a conformity score selected to achieve at least $1 - \delta$ coverage over a calibration set as in conformal prediction. CQR intervals offer the heterogeneous widths of quantile regression intervals as well as the marginal coverage guarantees of conformal prediction intervals.

Like conformal prediction, CQR has seen many uses and modifications. For instance, Hu et al.~\citeyearpar{wind_power_interval_prediction} combine CQR with temporal convolutional networks \citep{tcn} for the purpose of wind power interval prediction. Dietterich and Hostetler~\citeyearpar{mdp_prediction_intervals} extend CQR to multivariate responses to compute prediction intervals over the trajectories of MDP policies.

Let us now consider the ``backward problem''. Given a target interval, we want to find lower and upper values $\delta^-$ and $\delta^+$ such that the prediction interval produced by CQR, 
\[
\left[\hat{Q}_{Y|\vec{x}}\left(\delta^-\right) - c^-(\delta^-), \hat{Q}_{Y|\vec{x}}\left(1 - \delta^+\right) + c^+(\delta^+)\right],
\]
coincides with the target interval. Unfortunately, as we will show below, CQR is not invertible.  In this paper, we mitigate this issue by introducing a simple modification to CQR---we exploit the invertibility of the estimated conditional quantile function to move the conformity scores to the probability space. Our method, which we call Probability-space Conformalized Quantile Regression (PCQR), can be applied to the ``forward problem'' like CQR, but it can also be inverted and applied to the ``backward problem''.

We show that PCQR\textsuperscript{-1} can be viewed as an application of a conformally calibrated predictive systems \citep{conformal_calibrators} and, as such, enjoys the same conformal guarantees.

\section{Contributions}

This paper provides the following contributions:

\begin{enumerate}
    \item We show that even with an invertible estimated conditional quantile function, CQR itself is not invertible---following the quantile-space conformal correction, different estimated quantiles can result in the same prediction interval. Consequently, CQR cannot be employed to predict conditional coverage probabilities of user-specified target intervals.
    \item We introduce a simple modification to CQR, moving the conformal correction to the probability space by exploiting the invertibility of the estimated quantile function. We refer to our method as \textbf{P}robability-space \textbf{C}onformalized \textbf{Q}uantile \textbf{R}egression (PCQR).
    \item We show that by applying the conformal correction in the probability space, PCQR retains the invertibility of the estimated conditional quantile function. It can be inverted to estimate the CDF and predict conditional coverage probabilities of user-specified target intervals. We refer to this inverse algorithm as PCQR\textsuperscript{-1}. We show that PCQR\textsuperscript{-1} can also be viewed as an application of a conformally calibrated predictive system \citep{conformal_calibrators}.
    \item We prove that PCQR and PCQR\textsuperscript{-1} are well-calibrated and provide finite-sample marginal guarantees, and we evaluate both methods experimentally on two domains.
\end{enumerate}

\section{Method}

\subsection{CQR is not Invertible}

A natural question is whether CQR can be inverted to predict conditional coverage probabilities given a target interval. That is, is it possible to determine which conditional quantile estimates generate the target interval following the conformal correction? A necessary assumption is that the underlying quantile regression model is invertible. However, a simple counterexample shows that this assumption is insufficient. Given some conditioning input $\vec{x}$, suppose the quantile regression model estimates the conditional quantile function to be continuous and strictly increasing with respect to the cumulative probability, with
\begin{equation*}
    \begin{aligned}
        \hat{Q}_{Y|\vec{x}}(0.05) & = a,\\
        \hat{Q}_{Y|\vec{x}}(0.1) & = a + \epsilon,\\
        \hat{Q}_{Y|\vec{x}}(0.9) & = b - \epsilon,\\
        \hat{Q}_{Y|\vec{x}}(0.95) & = b,
    \end{aligned}
\end{equation*}
\noindent for some $a, b, \epsilon \in \mathbf{R}$, such that
\begin{equation*}
    \begin{aligned}
        a & < b, \quad\mbox{and}\\
        0 < \epsilon & < \frac{b - a}{2}.
    \end{aligned}
\end{equation*}
\noindent Suppose that on the calibration set, the underlying quantile regression model suffers from overcoverage for $90\%$ prediction intervals and undercoverage for $80\%$ prediction intervals. Then the conformal correction $c_{0.1}$ (as in \refeq{conformalized_quantile_regression}) will be negative, and the conformal correction $c_{0.2}$ will be positive. Suppose that $c_{0.2} - c_{0.1} = \epsilon$. Then the CQR $80\%$ prediction interval is
\begin{equation*}
    \begin{aligned}
        CQR(\vec{x}, 0.2) & = [a + \epsilon - c_{0.2}, b - \epsilon + c_{0.2}]\\
        & = [a - c_{0.1}, b + c_{0.1}]\\
        & = CQR(\vec{x}, 0.1),
    \end{aligned}
\end{equation*}
\noindent which is also the CQR $90\%$ prediction interval. That is, the quantile-space conformal correction destroys the invertibility of the conditional quantile estimates. Similar counterexamples show that CQR can introduce quantile crossings, even when applied on top of a strictly increasing estimate of the conditional quantile function.

\subsection{Introducing Probability-Space Conformity Scores}

To solve the invertibility issue and predict coverage probabilities given target intervals, we start by making a simple modification to CQR. Our method, like CQR, constructs intervals by applying a conformal correction to an initial estimate of the conditional quantile function. For the rest of the paper, we will assume that, as in the counterexample, the estimated conditional quantile function $\hat{Q}_{Y|\vec{x}}(\alpha)$ is continuous and strictly increasing with respect to the cumulative probability $\alpha$.

CQR applies a per-$\alpha$ additive conformal correction to the estimated quantile. While $\hat{Q}_{Y|\vec{x}}(\alpha)$ is a strictly increasing function of $\alpha$, the correction is not, and the corrected quantile estimate is not necessarily invertible. We mitigate this issue by exploiting the invertibility of the estimated conditional quantile function and moving the conformal correction to the probability space. That is, our method predicts intervals of the form
\begin{equation}
    PCQR(\vec{x}, \delta) = \left[\hat{Q}_{Y|\vec{x}}\left(\frac{\delta}{2} - c_{\delta}\right), \hat{Q}_{Y|\vec{x}}\left(1 - \frac{\delta}{2} + c_{\delta}\right)\right],
\end{equation}
\noindent where $c_{\delta}$ is chosen to achieve coverage of at least $1 - \delta$ on the calibration set, as in conformal prediction and CQR. Because the conformal correction is applied to the cumulative probability rather than the estimated conditional quantile, we refer to our method as \textbf{P}robability-space \textbf{C}onformalized \textbf{Q}uantile \textbf{R}egression (PCQR).

CQR corrects the estimated conditional quantiles associated with a $1 - \delta$ coverage rate in order to achieve at least $1 - \delta$ coverage on the calibration set. In contrast, PCQR determines \textit{which} estimated conditional quantiles achieve at least $1 - \delta$ coverage on the calibration set without any conformal corrections. In both methods, the chosen cumulative probabilities are symmetric about $\frac{1}{2}$. Additionally, the conformal corrections in both methods can be modified to predict an interval whose marginal coverage probability is \textit{at most} $1 - \delta$, rather than at least $1 - \delta$. In light of this interpretation, PCQR can instead be expressed as
\begin{equation}
    \label{eq:pcqr}
    PCQR(\vec{x}, s) = \left[\hat{Q}_{Y|\vec{x}}\left(\frac{1}{2} - s\right), \hat{Q}_{Y|\vec{x}}\left(\frac{1}{2} + s\right)\right],
\end{equation}
\noindent where $s$ is the calibration conformity score selected to achieve the desired marginal coverage bound.

To compute $s$, we leverage the same techniques used in split conformal prediction and CQR. Let \mbox{$D_{cal} = \{(\vec{x_i}, y_i)\}_{i=1}^n$} denote the calibration set, where $\vec{x_i}$ is the feature vector (i.e., of predictor variables), and $y_i$ is the scalar response variable. Let $\vec{x_{n+1}}$ denote a new feature vector observed at test time, and let $y_{n+1}$ denote the corresponding unobserved response variable. Assume that neither $(\vec{x_{n+1}}, y_{n+1})$ nor any of the data in $D_{cal}$ have been observed by the quantile regression model during training. Assume that the random sequence $(\vec{x_1}, y_1), \ldots, (\vec{x_{n+1}}, y_{n+1})$ is exchangeable (i.e., the calibration and testing data are exchangeable). Let $\hat{F}_{Y|\vec{x}}$ denote the inverse of $\hat{Q}_{Y|\vec{x}}$, so that
\begin{equation}
    \label{eq:f_is_inverse_q}
    \hat{F}_{Y|\vec{x}}(y) = \alpha \iff \hat{Q}_{Y|\vec{x}}(\alpha) = y.
\end{equation}
\noindent Given that $\hat{Q}_{Y|\vec{x}}$ is a continuous and strictly increasing estimate of the conditional quantile function, $\hat{F}_{Y|\vec{x}}$ serves as a continuous and strictly increasing estimate of the conditional CDF.

The conformity score for a given instance $(\vec{x}, y)$ is computed as
\begin{equation}
    \label{eq:s}
    S(\vec{x}, y) = \Big|\frac{1}{2} - \hat{F}_{Y|\vec{x}}(y)\Big|,
\end{equation}
\noindent where $|\cdot|$ denotes absolute value.

Let $S_{(i)}$ denote the $i^{th}$ order statistic of the sequence of calibration conformity scores, $\{S(\vec{x_i}, y_i)\}_{i=1}^n$, and assume that they are almost surely unique (i.e., unique with probability $1$). We use these conformity scores to construct two PCQR interval predictions:
\begin{align}
    I^-(\vec{x_{n+1}}, \delta) & = PCQR(\vec{x_{n+1}}, S_{(\lfloor (1 - \delta) (n + 1) \rfloor)})\\
    I^+(\vec{x_{n+1}}, \delta) & = PCQR(\vec{x_{n+1}}, S_{(\lceil (1 - \delta) (n + 1) \rceil)})
\end{align}
\noindent By construction, \mbox{$I^+(\vec{x_{n+1}}, \delta)$} is only defined for \mbox{$\delta \geq \frac{1}{n+1}$}, and \mbox{$I^-(\vec{x_{n+1}}, \delta)$} is only defined for \mbox{$\delta > 0$}. Similar constraints apply to CQR prediction intervals.

\begin{theorem}
    \label{thm:pcqr}
    Given an exchangeable sequence $(\vec{x_1}, y_1), \ldots, (\vec{x_{n+1}}, y_{n+1})$ and almost surely unique conformity scores $S(\vec{x_1}, y_1), \ldots, S(\vec{x_{n+1}}, y_{n+1})$, the following inequalities hold:
        \begin{equation}
            P(y_{n+1} \in I^-(\vec{x_{n+1}}, \delta))
            \leq 1 - \delta
            \leq P(y_{n+1} \in I^+(\vec{x_{n+1}}, \delta)).
    \end{equation}
    
    \begin{proof}
        Because the tuples $(\vec{x_1}, y_1), \ldots, (\vec{x_{n+1}}, y_{n+1})$ are exchangeable, and because each conformity score $S(\vec{x_i}, y_i)$ is a function $S(\cdot)$ of the corresponding tuple $(\vec{x_i}, y_i)$, the conformity scores $S(\vec{x_1}, y_1), \ldots, S(\vec{x_{n+1}}, y_{n+1})$ are also exchangeable.
        
        The rank of any variable in an exchangeable sequence of almost surely unique random variables is uniformly distributed. Consequently, the marginal distribution of the rank of each conformity score is uniform in \mbox{$\{1, 2, \ldots, n+1\}$}.
        
        Then we have that,
        \begin{equation}
            \begin{aligned}
                P(S(\vec{x_{n+1}}, y_{n+1}) \leq S_{(i)}) & = P(R(S(\vec{x_{n+1}}, y_{n+1})) \leq i)\\
                & = \frac{i}{n+1},
            \end{aligned}
        \end{equation}
        \noindent where $R(\cdot)$ denotes rank in the sequence of all $n+1$ conformity scores. We can apply this to compute coverage probability bounds on the selected conformity scores:
        \begin{align}
            \label{eq:lower_bound_proof}
            & \begin{aligned}
                P(S(\vec{x_{n+1}}, y_{n+1}) \leq S_{(\lfloor (1 - \delta) (n + 1) \rfloor)}) & = \frac{\lfloor (1 - \delta) (n + 1) \rfloor}{n + 1}\\
                & \leq 1 - \delta
            \end{aligned}\\
            \label{eq:upper_bound_proof}
            & \begin{aligned}
                P(S(\vec{x_{n+1}}, y_{n+1}) \leq S_{(\lceil (1 - \delta) (n + 1) \rceil)}) & = \frac{\lceil (1 - \delta) (n + 1) \rceil}{n + 1}\\
                & \geq 1 - \delta.
            \end{aligned}
        \end{align}
        
        Recall that $\hat{Q}_{Y|\vec{x}}$ and $\hat{F}_{Y|\vec{x}}$ are assumed to be continuous and strictly increasing. Then the following logical biconditionals hold:
        \begin{equation}
            \label{eq:inverse_coverage_implications}
            \begin{aligned}
                & y \in PCQR(\vec{x}, s)\\
                & \hspace{20pt} \iff \hat{Q}_{Y|\vec{x}}\Big(\frac{1}{2} - s\Big) \leq y \leq \hat{Q}_{Y|\vec{x}}\Big(\frac{1}{2} + s\Big)\\
                & \hspace{20pt} \iff \frac{1}{2} - s \leq \hat{F}_{Y|\vec{x}}(y) \leq \frac{1}{2} + s\\
                & \hspace{20pt} \iff \Big| \frac{1}{2} - \hat{F}_{Y|\vec{x}}(y) \Big| \leq s\\
                & \hspace{20pt} \iff S(\vec{x}, y) \leq s.
            \end{aligned}
        \end{equation}
        
        Combining \refeq{lower_bound_proof}, \refeq{upper_bound_proof} and \refeq{inverse_coverage_implications} yields the following results:
        \begin{align}
            & \begin{aligned}
                & P(y_{n+1} \in I^-(\vec{x_{n+1}}, \delta))\\
                & \hspace{20pt} = P(y_{n+1} \in PCQR(\vec{x_{n+1}}, S_{(\lfloor (1 - \delta) (n + 1) \rfloor)}))\\
                & \hspace{20pt} = P(S(\vec{x_{n+1}}, y_{n+1}) \leq S_{(\lfloor (1 - \delta) (n + 1) \rfloor)})\\
                & \hspace{20pt} \leq 1 - \delta
            \end{aligned}\\
            & \begin{aligned}
                & P(y_{n+1} \in I^+(\vec{x_{n+1}}, \delta))\\
                & \hspace{20pt} = P(y_{n+1} \in PCQR(\vec{x_{n+1}}, S_{(\lceil (1 - \delta) (n + 1) \rceil)}))\\
                & \hspace{20pt} = P(S(\vec{x_{n+1}}, y_{n+1}) \leq S_{(\lceil (1 - \delta) (n + 1) \rceil)})\\
                & \hspace{20pt} \geq 1 - \delta.
            \end{aligned}
        \end{align}
    \end{proof}
\end{theorem}

\refthm{pcqr} shows that the marginal coverage probability of $I^+(\vec{x_{n+1}}, \delta)$ is at least $1 - \delta$. Hence, it is a marginally valid prediction interval. Since the coverage probabilities of $I^-(\vec{x_{n+1}}, \delta)$ and $I^+(\vec{x_{n+1}}, \delta)$ are separated by at most $\frac{1}{n+1}$, the prediction intervals are also efficient. It is important to note that \refthm{pcqr} does not provide conditional coverage guarantees for any fixed $\vec{x_{n+1}}$ or calibration set; it only provides marginal coverage guarantees over all possible random sequences of calibration and test data, given that the data from the two sets are exchangeable. These are the same guarantees provided by CQR.

\subsection{PCQR is Invertible}

Following the expression of PCQR in \refeq{pcqr}, it is easy to see that the algorithm can be inverted to compute the conditional coverage probability within a user-specified target interval. Indeed, the upper and lower bounds of a PCQR prediction interval are each directly computed by the estimated conditional quantile function. The estimated conditional CDF, then, can be employed to compute the conformally-corrected cumulative probabilities associated with the target interval bounds. These cumulative probabilities can be used to measure coverage over the calibration set.

Let $[y^-, y^+]$ denote the user-specified target interval. Let $\alpha_i = \hat{F}_{Y|\vec{x_i}}(y_i)$, and let $\alpha_{(i)}$ denote the $i^{th}$ order statistic of the sequence of calibration cumulative probabilities, $\{\alpha_i\}_{i=1}^n$.

\begin{definition}
    We define a conformally calibrated conditional CDF estimate as follows:
    \begin{equation}
        F_{Y|\vec{x_{n+1}}}^{cal}(y) = \frac{\max \{i : \alpha_{(i)} < \hat{F}_{Y|\vec{x_{n+1}}}(y)\}}{n + 1} + \frac{1}{2n + 2}.    
\end{equation}
\end{definition}

We then define $p^-(\cdot)$ and $p^+(\cdot)$, which we refer to as the PCQR\textsuperscript{-1} probability lower bound and PCQR\textsuperscript{-1} probability upper bound, respectively:
\begin{align}
    p^-(\vec{x_{n+1}}, y^-, y^+) = F_{Y|\vec{x_{n+1}}}^{cal}(y^+) - F_{Y|\vec{x_{n+1}}}^{cal}(y^-) - \frac{1}{n+1}\\
    p^+(\vec{x_{n+1}}, y^-, y^+) = F_{Y|\vec{x_{n+1}}}^{cal}(y^+) - F_{Y|\vec{x_{n+1}}}^{cal}(y^-) + \frac{1}{n+1}
\end{align}

$F_{Y|\vec{x_{n+1}}}^{cal}(y)$ is a conformally calibrated conditional CDF estimate, so PCQR\textsuperscript{-1} can be viewed as an application of the work of \citet{conformal_calibrators} on conformally calibrated predictive systems.

\begin{lemma}
    \label{lemma:uniform_calibration}
    Let $Z$ be a discrete random variable uniform on $\{\frac{1}{2k}, \frac{3}{2k}, ..., \frac{2k-1}{2k}\}$ for some positive integer $k$, and let $F_Z$ be the CDF of $Z$. Then for all $z\in [0,1]$ and all $k$,

    \begin{equation}
            z - \frac{1}{2k} < F_Z(z) \leq z + \frac{1}{2k}.
            \label{eq:uniform_calibration}
    \end{equation}

    \begin{proof}
        Let $s_i = \frac{2i-1}{2k}$. Let $S = \{\frac{1}{2k}, \frac{3}{2k}, ..., \frac{2k-1}{2k}\} = \{s_i\}_{i=1}^{k}$ denote the support of $Z$.

        Applying the definition of a discrete uniform CDF, we can compute $F_Z(s_i)$ for any $s_i \in S$ directly:
        \begin{equation}
            \label{eq:cdf_si}
            \begin{aligned}
                F_Z(s_i) & = P(Z \leq s_i)\\
                & = \frac{\sum_{j \in \{1, 2, ..., k\}} \mathbbm{1}_{(-\infty, s_i]}(s_j)}{k}\\
                & = \frac{i}{k}\\
                & = \frac{2i-1}{2k} + \frac{1}{2k}\\
                & = s_i + \frac{1}{2k}.
            \end{aligned}
        \end{equation}

        We divide the proof into three cases. 

        Case 1: Consider the left-most interval where $z \in [0,1/2k)$. In this interval, $z-1/2k\in [-1/2k,0)$, $z+1/2k\in[1/2k,1/k)$, and $F_Z(z)=0$. Hence, (\ref{eq:uniform_calibration}) is satisfied. 
        
        Case 2: Now consider the right-most interval where $z \in ((2k-1)/2k,1]$. Here, $z-1/2k \in (\frac{2k-2}{2k},\frac{2k-1}{2k}]$, $z+1/2k\in (1, \frac{2k+1}{2k}]$, and $F_Z(z)=1$. Again (\ref{eq:uniform_calibration}) is satisfied. 
        
        Case 3: Finally, suppose $\frac{1}{2k} \leq z \leq \frac{2k-1}{2k}$. Let $s^*(z) = \max\{s_i \in S : s_i \leq z\}$. Because each adjacent $s_i$ is separated by a constant interval $\frac{1}{k}$

        \begin{equation}
        \label{eq:s_star_bounds}
            z - \frac{1}{k} < s^*(z) \leq z.
        \end{equation}

        We can combine the definition of $F_Z(z)$ with \refeq{cdf_si} and \refeq{s_star_bounds} to complete the proof:
        \begin{equation}
            \begin{aligned}
                F_Z(z) & = P(Z \leq z)\\
                & = P(Z \leq s^*(z)) + P(s^*(z) < Z \leq z)\\
                & = P(Z \leq s^*(z))\\
                & = s^*(z) + \frac{1}{2k}\\
                & > \Big(z - \frac{1}{k}\Big) + \frac{1}{2k}\\
                & = z - \frac{1}{2k},
            \end{aligned}
        \end{equation}

        and similarly,

        \begin{equation}
            \begin{aligned}
                F_Z(z) & = ...\\
                & = s^*(z) + \frac{1}{2k}\\
                & \leq z + \frac{1}{2k}.
            \end{aligned}
        \end{equation}
    \end{proof}

\end{lemma}

\begin{theorem}
    \label{thm:pcqr_inverse}
    Given an exchangeable sequence $(\vec{x_1}, y_1), \ldots, (\vec{x_{n+1}}, y_{n+1})$ and almost surely unique conditional cumulative probability estimates $\{\alpha_1, \ldots, \alpha_{n+1}\}$, $F_{Y|\vec{x_{n+1}}}^{cal}(y_{n+1})$ is a well-calibrated CDF estimate
    \begin{equation}
        \label{eq:inverse_pcqr_thm}
               \beta - \frac{1}{2n + 2} \;\; < \;\;P(F_{Y|\vec{x_{n+1}}}^{cal}(y_{n+1}) \leq \beta) \;\; \leq \;\;\beta + \frac{1}{2n + 2}
     \end{equation}

    for all $\beta \in [0, 1]$.
    
    \begin{proof}
        The standard conformal argument applies: $(\vec{x_1}, y_1), \ldots, (\vec{x_{n+1}}, y_{n+1})$ are exchangeable, and each $\alpha_i$ is a function $\hat{F}_{Y|\vec{x_i}}(\cdot)$ of the corresponding $y_i$. Hence, the estimated conditional cumulative probabilities $\alpha_1, \ldots, \alpha_{n+1}$ are also exchangeable. Given that each $\alpha_i$ is almost surely unique, the marginal distribution of the rank of each $\alpha_i$ is uniform in $\{1, 2, \ldots, n+1\}$.

        We can rewrite $F_{Y|\vec{x_{n+1}}}^{cal}(y_{n+1})$ in terms of ranks,

        \begin{equation}
            \label{eq:f_cal_rank}
            \begin{aligned}
                F_{Y|\vec{x_{n+1}}}^{cal}(y_{n+1}) & = \frac{\max \{i : \alpha_{(i)} < \hat{F}_{Y|\vec{x_{n+1}}}(y_{n+1})\}}{n + 1} + \frac{1}{2n + 2}\\
                & = \frac{R(\alpha_{n+1}) - 1}{n+1} + \frac{1}{2n + 2},
            \end{aligned}
        \end{equation}

        where $R(\alpha_{n+1})$ denotes the rank of $\alpha_{n+1}$ in the sequence $\alpha_1, ..., \alpha_{n+1}$. Because the marginal distribution of $R(\alpha_{n+1})$ is uniform with support $\{1, ..., n+1\}$, the marginal distribution of $F_{Y|\vec{x_{n+1}}}^{cal}(y_{n+1})$ is uniform with support $\{\frac{1}{2n + 2}, \frac{3}{2n+2}, ..., \frac{2n + 1}{2n + 2}\}$.

        Therefore, we can apply \reflemma{uniform_calibration} to complete the proof, substituting $Z = F_{Y|\vec{x_{n+1}}}^{cal}(y_{n+1})$, $k=n+1$, and $z=\beta$.
    \end{proof}
\end{theorem}

Hence, the coverage probabilities $p^-(\vec{x_{n+1}}, y^-, y^+)$ and $p^+(\vec{x_{n+1}}, y^-, y^+)$ are each computed as the difference between the calibrated estimates of the conditional CDF of $y_{n+1} \mid \vec{x}_{n+1}$ evaluated at the upper and lower interval bounds, corrected to account for the $\frac{1}{n+1}$ calibration error. Like traditional conformal guarantees, the calibration guarantee provided by \refthm{pcqr_inverse} is purely marginal---it does not hold when conditioning on any specific calibration points, test points, or user-specified interval bounds. Nevertheless, it is a necessary guarantee for any true CDF.

\section{Experiments}

We evaluate PCQR and PCQR\textsuperscript{-1} on two MDPs: \textit{Starcraft 2} \citep{starcraft} and \textit{Tamarisk} \citep{tamarisk}. In each domain, we sample a total of $n=10{,}000$ episodes. We then randomly partition the episodes into $D_{train}$ with $n_{train}=2{,}500$ episodes, $D_{cal}$ with $n_{cal}=2{,}500$ episodes, and $D_{test}$ with $n_{test}=5{,}000$ episodes for testing. We repeat each experiment with $k=5$ different random partitions of the data. For a fixed episode and a given time step $t$, the feature vector $\vec{x_t}$ is a subset of the MDP state variables, and the response variable $y_t$ is the total reward at the end of the episode. The target interval is determined by computing a marginal $80\%$ interval $[y^-, y^+]$ about the final cumulative reward for all episodes in $D_{test}$ using the empirical $0.1$- and $0.9$-quantiles. 

Starcraft 2 is a small subset of the famous RTS game. In this subset, a team of blue units faces off against a team of red units. In this particular evaluation, the initial number of blue units is chosen uniformly in $\{5, \ldots, 20\}$, and the initial number of red units is chosen uniformly in $\{5, \ldots, 10\}$. The blue team is controlled by a fixed policy commanding its units to advance toward the red team at each time step. When the teams meet, they engage in battle under the built-in logic of Starcraft 2.  At time step $t=14$, the red team receives a random number of reinforcements chosen uniformly in ${0, \ldots, N}$, where $N$ is chosen uniformly in ${0, \ldots, 15}$. The blue team receives $+1$ reward for each eliminated red unit and $-1$ reward for each eliminated blue unit. To avoid ties in PCQR conformity scores and estimated conditional cumulative probabilities, we add uniform random noise in $[-5\times 10^{-6}, 5\times 10^{-6}]$ to the reward at the final time step ($H=57$) of each episode.

The Tamarisk domain involves controlling the plant species present within a river network. The river network is represented by a balanced binary tree with seven edges. At any given time, each edge is either empty, occupied by a native plant, or occupied by an invasive tamarisk tree. The initial state of each edge is chosen uniformly. The agent's action is a vector of seven primitive actions, each corresponding to an edge in the river network. The available primitive actions are 1) do nothing; 2) eradicate the existing tamarisk tree, provided that one is present; 3) plant a native plant, provided that the edge is empty; and 4) eradicate the existing tamarisk tree, followed by planting a native plant. Each primitive action is associated with a cost, and there is a budget on the sum of the seven primitive action costs at each time step. Additionally, there is a cost at each time step for each edge occupied by a tamarisk tree. The reward is computed as the negative cost, so the reward at each time step is non-positive. We implemented a fixed policy based on the work of \citet{tamarisk}.
As in the Starcraft 2 domain, we add uniform random noise in $[-5\times 10^{-6}, 5\times 10^{-6}]$ to the reward at the final time step ($H=50$) of each episode to avoid ties in PCQR conformity scores and estimated conditional cumulative probabilities.

In each domain, for each time step $t$, we train a quantile regression forest (QRF) on the data in $D_{train}$ corresponding to time step $t$ to estimate the quantile function of $y_t$ conditioned on $\vec{x_t}$ (i.e., $\hat{Q}_{Y_t|\vec{x_t}}$). QRFs estimate conditional quantiles as weighted empirical quantiles over the response variables in the training data. We inverted them to estimate the conditional CDF (i.e., $\hat{F}_{Y_t|\vec{x_t}}$) by replacing the weighted empirical quantile by the weighted empirical CDF.

To evaluate the performance of PCQR, we apply it to the data in $D_{cal}$ corresponding to time step $t{=}0$ using the fitted QRFs. For each episode in $D_{test}$, we predict marginally valid $80\%$ PCQR prediction intervals (i.e., $I^+(\vec{x_0}, 0.2)$) for $y_H$ conditioned on $\vec{x_0}$. We measured the mean and standard deviation of the empirical coverage of these prediction intervals across all $k$ data partitions to be $80.32\% \pm 1.07\%$ for Starcraft 2 and $79.89\% \pm 0.79\%$ for Tamarisk; this provides evidence that PCQR is well-calibrated for these domains.

To evaluate the performance of PCQR\textsuperscript{-1}, we apply PCQR\textsuperscript{-1} to the data in $D_{cal}$ separately for each time step $t$. For each episode, we translate the target interval $[y^-, y^+]$ to the space of reward-to-go values for time step $t$ by subtracting off the cumulative reward observed up to time step $t$. We then apply PCQR\textsuperscript{-1} to this translated interval to predict lower and upper bounds on the probability that the final cumulative reward will fall in $[y^-, y^+]$. This involves fitting conditional CDF models to predict the CDF of the reward-to-go from $t$ to the end of the episode ($H$).

\begin{figure}[htb!]
    \centering
    \includegraphics[width=0.5\columnwidth]{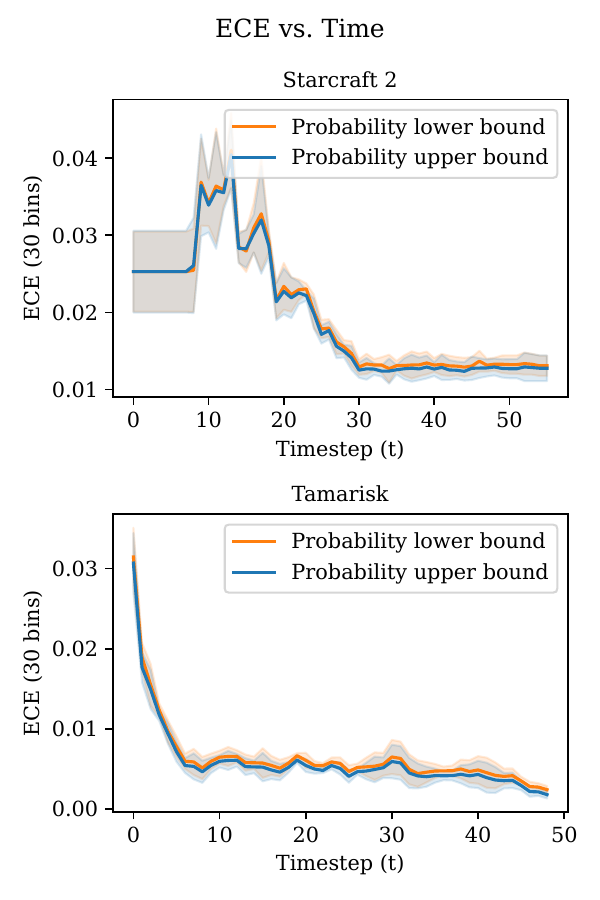}
    \caption{Mean Expected Calibration Error (ECE) across all $k$ data partitions vs. time for the PCQR\textsuperscript{-1} probability lower and upper bound predictions in the Starcraft 2 and Tamarisk domains. ECE is measured with 30 equal-width bins. Standard deviations are depicted by the semi-transparent regions.}
    \label{fig:ece_vs_time}
\end{figure}

\begin{figure}[htb!]
    \centering
    \includegraphics[width=0.5\columnwidth]{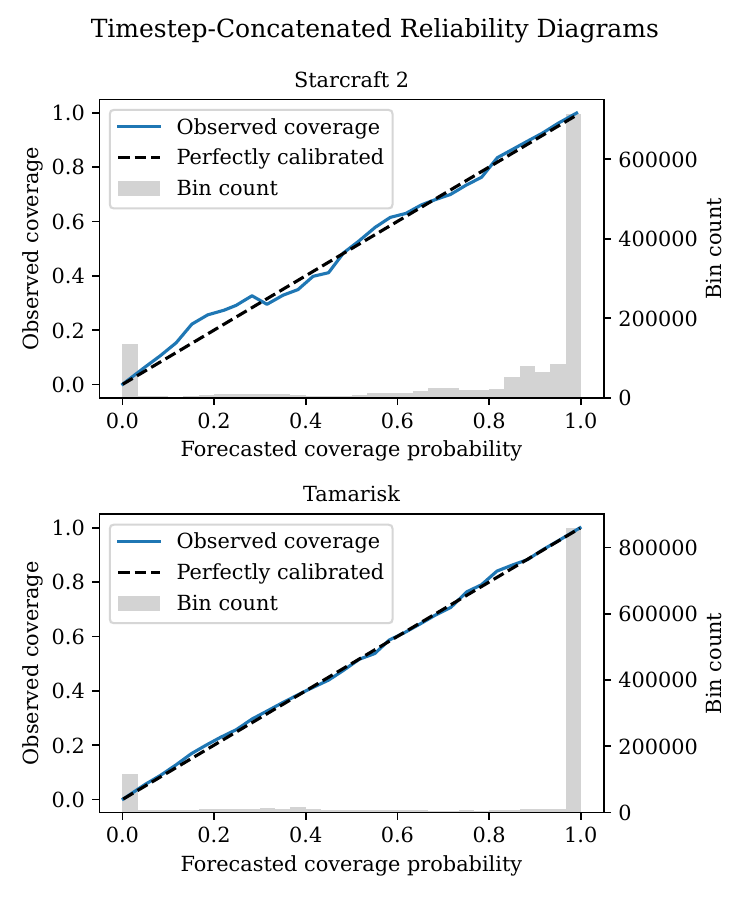}
    \caption{Reliability diagrams for the PCQR\textsuperscript{-1} probability lower bound predictions in the Starcraft 2 and Tamarisk domains. Data from all data partitions and time steps are binned into 30 equal-width bins according to the predicted coverage probabilities. The gray histograms give the bin counts. The x-axis depicts the mean predicted coverage probability in each bin. The y-axis shows the observed coverage rate in each bin. The blue line corresponds to the experimental results, and the black dashed line reflects perfect calibration.}
    \label{fig:reliability_diagrams}
\end{figure}

\reffig{ece_vs_time} plots the mean Expected Calibration Error \cite[ECE;][]{Naeini2015} of the PCQR\textsuperscript{-1} probability bounds across all $k$ data partitions as a function of time. We measured ECE with 30 equal-width bins. Additionally, \reffig{reliability_diagrams} visualizes reliability diagrams for PCQR\textsuperscript{-1} probability lower bounds by combining the predictions across all time steps and all $k$ data partitions. The x-axis depicts the mean predicted coverage probability in each bin, and the y-axis depicts the bin's observed empirical coverage rate. Perfect calibration is reflected by an ECE of zero. In a reliability diagram, perfect calibration is reflected by a diagonal line along $y=x$. As shown in the figures, the ECE is relatively low, and the reliability diagram curves are close to the diagonal. Lastly, we combine the predictions across all time steps to compute a single mean and standard deviation of the ECE for PCQR\textsuperscript{-1} probability lower bounds across all $k$ data partitions. These values were measured as $0.013 \pm 0.0018$ for Starcraft 2 and $0.004 \pm 0.0005$ for Tamarisk. These results provide evidence that PCQR\textsuperscript{-1} is well-calibrated for these domains.

\begin{figure}[htb!]
    \centering
    \includegraphics[width=0.5\columnwidth]{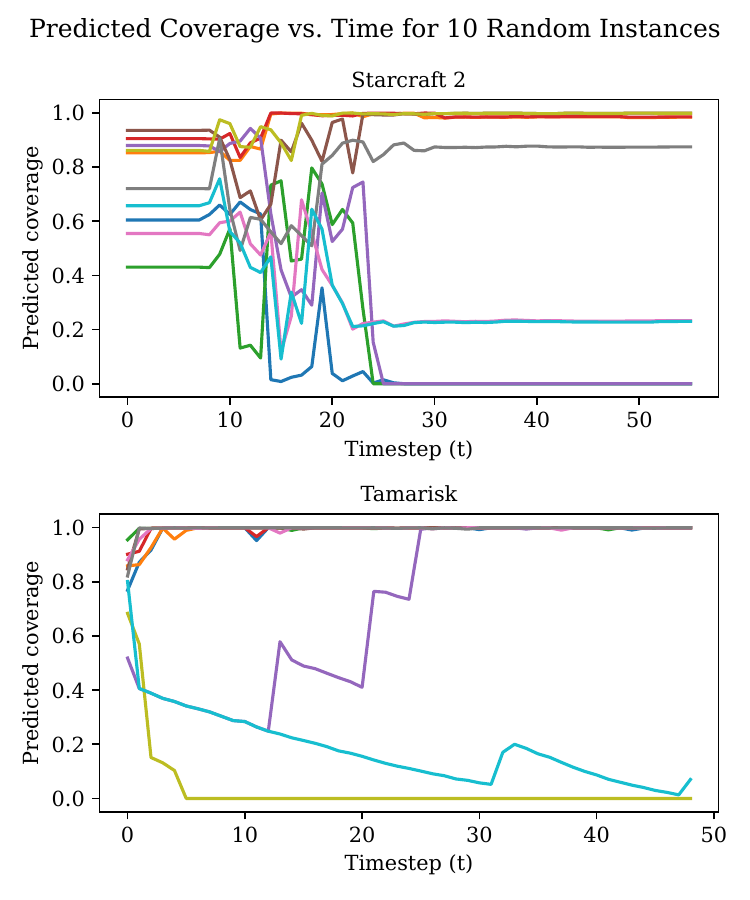}
    \caption{PCQR\textsuperscript{-1} predicted coverage probability lower bounds vs.\ time for 10 random episodes in the Starcraft 2 and Tamarisk domains. Each line corresponds to a separate episode.}
    \label{fig:10_instances_predicted_coverage}
\end{figure}

To better understand the behavior of PCQR\textsuperscript{-1}, we plot the predicted PCQR\textsuperscript{-1} probability lower bound as a function of time for 10 randomly-selected episodes from a single random data partition; these plots are depicted in \reffig{10_instances_predicted_coverage}. Many of the predicted coverage probabilities converge to either $0$ or $1$ by the penultimate time step of the episode. This is because, by the end of the episode, there is little uncertainty as to whether the final cumulative reward will land in the target interval. Importantly, however, some predicted coverage probabilities do \textit{not} converge. This is because there are duplicated response values in the original datasets prior to adding uniform random noise. After applying noise, the distribution of response variables is often multimodal, with a mode in a $\pm 5\times 10^{-6}$ interval around each duplicated response value. The distribution of response variables within each mode is uniform, matching the distribution of the added noise. When the target interval boundaries fall within one of these modes, any episodes with response variables falling within the same mode will have a constant amount of aleatoric uncertainty until the final time step, at which point the noise is observed and the final cumulative reward is known. The quantity of this aleatoric uncertainty depends on the location of the interval boundary within the mode. In other words, this pattern is merely an artifact of the artificial noise added to break ties in conformity scores and estimated conditional cumulative probabilities, along with the empirical construction of the target interval. In practice, the interval bounds could be expanded on both sides by $10^{-5}$ (an amount equal to the range of the tie-breaking noise perturbations) to include all of the values in the mode, causing all of the unconverged probabilities to converge to $1$. Likewise, the interval bounds could be contracted on both sides by $10^{-5}$ to cause them to converge to $0$. In some sense, expanding the interval bounds treats it like a closed interval, and contracting the interval bounds treats it like an open interval. The best choice is application-dependent.

Additionally, some of the curves for the Tamarisk domain show intermittent spikes in PCQR\textsuperscript{-1} predictions throughout the episode. While the cumulative reward in the Tamarisk domain converges in fewer than 5 time steps for most episodes, there is a small probability that each plant will die a natural death in each time step. Once the invasive tamarisk plants have been eliminated, the river network edges are all occupied by native plants. Hence, a native plant death will cause a small decrease in the cumulative reward. Following the death, the agent quickly plants a new native plant to fill the empty plot, and the cumulative reward stabilizes. If the episode's cumulative reward is close to one of the interval bounds, such plant deaths can cause sudden changes in PCQR\textsuperscript{-1} predictions, as we observe in \reffig{10_instances_predicted_coverage}.

\begin{figure}[htb!]
    \centering
    \includegraphics[width=0.5\columnwidth]{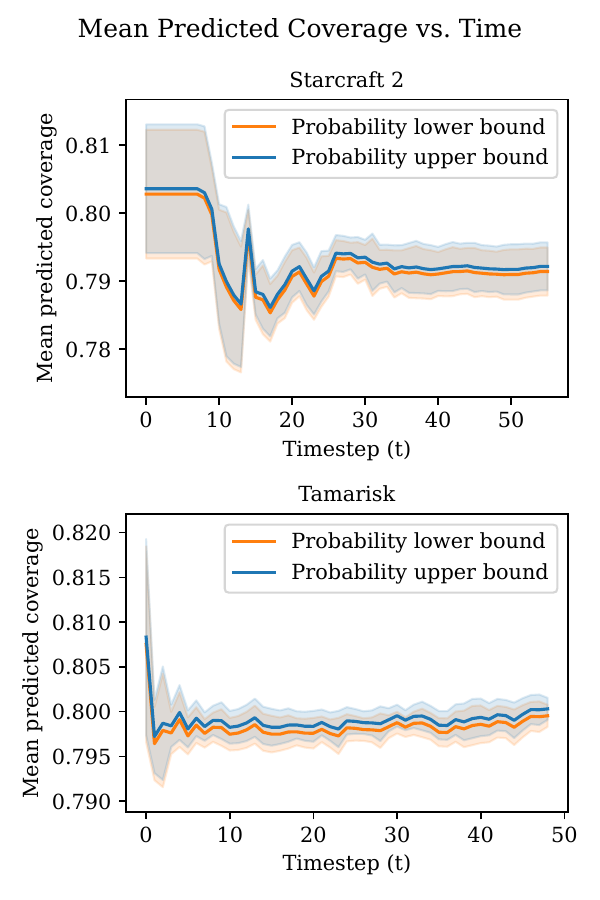}
    \caption{Mean PCQR\textsuperscript{-1} predictions vs. time in the Starcraft 2 and Tamarisk domains. Means are computed across all episodes of all data partitions. Standard deviations are computed from the per-data-partition means across all episodes and are depicted by the semi-transparent regions.}
    \label{fig:mean_predicted_coverage}
\end{figure}

Finally, to summarize the performance of PCQR\textsuperscript{-1} for each domain, we plot the mean predicted conditional coverage probability across all episodes and all $k$ data partitions as a function of time; these plots are depicted in \reffig{mean_predicted_coverage}. Although the predicted conditional coverage probability varies over time for each episode, the mean predicted conditional coverage probability across all episodes is close to $0.8$ for every time step.

\section{Conclusion}

Previous work has focused on conditional prediction intervals, which are of great interest in applications. However, there are other scenarios where the closely related task of conditional interval coverage prediction is also important. By introducing a simple modification to conformalized quantile regression, we are able to construct an algorithm that can easily be inverted to provide an elegant solution for both tasks.

Both tasks are important in the discussion of reliable autonomy. Conditional prediction intervals can provide important insight into the future behavior of an autonomous system in a stochastic environment, provided that those intervals are efficient and valid. Conditional interval coverage prediction can provide interpretable estimates for the probability that an autonomous system's performance will fall inside a user-specified target interval, provided that the those estimates are well-calibrated.

\subsection{Future work}

The experiments in this paper were conducted entirely with quantile regression forests serving as the underlying estimate of the conditional quantile function. Further experiments could study the performance of PCQR and PCQR\textsuperscript{-1} under other quantile regression methods, such linear quantile regression \cite{regression_quantiles} and implicit quantile networks \cite{implicit_quantile_networks}.

While the focus of this paper was PCQR\textsuperscript{-1} and the task of conditional interval coverage prediction, PCQR serves as a interesting alternative to CQR, conformal prediction, and quantile regression. Further experiments should compare the relative performance of all of the aforementioned methods. Moreover, given that PCQR\textsuperscript{-1} exploits the invertibility of the underlying quantile regression model, a simple baseline would be to use the conditional cumulative probability estimates from the inverted quantile regression to directly predict the conditional interval coverage probability, in a manner similar to quantile regression without conformalization. Further experiments are required to compare PCQR\textsuperscript{-1} against this baseline.

Finally, this paper describes PCQR and PCQR\textsuperscript{-1} in the context of univariate responses. In many applications, the response variable of interest is multivariate. \citet{mdp_prediction_intervals} extend CQR to form prediction intervals about the entire trajectory of the cumulative reward achieved by MDP policies; a natural next step would be to extend PCQR and PCQR\textsuperscript{-1} to multivariate responses as well. It would be nice to find a method that could guarantee the simultaneous calibrated accuracy of \textit{all} of the probabilities predicted by PCQR\textsuperscript{-1} for $t=0,\ldots, H-1$, rather than the single-time-step guarantees provided in this paper.

\subsection{Acknowledgments}
This material is based upon work supported by the Defense Advanced Research Projects Agency (DARPA) under Contract No. HR001119C0112. Any opinions, findings and conclusions or recommendations expressed in this material are those of the author(s) and do not necessarily reflect the views of the DARPA. The authors thank Jesse Hostetler for providing the Starcraft 2 trajectories and an anonymous reviewer for pointing out the connection to \cite{conformal_calibrators}.

\bibliographystyle{plainnat}
\bibliography{main}

\end{document}